\documentclass{article}
\usepackage{spconf,amsmath,graphicx}
\usepackage{booktabs}
\usepackage{adjustbox}
\usepackage{times}
\usepackage{epsfig}
\usepackage{amsmath}
\usepackage{amssymb}
\usepackage{color}
\usepackage{colortbl}
\usepackage{pifont}
\usepackage{subfigure}
\usepackage{multirow}
\usepackage{algorithm}
\usepackage{algpseudocode}

\newcommand{\ie}{\textit{i}.\textit{e}.}
\newcommand{\etal}{\textit{et al}.}

\newcommand{\figref}[1]{Fig. \ref{#1}}
\newcommand{\tabref}[1]{Table \ref{#1}}

\newcommand{\secref}[1]{Sec. \ref{#1}}
\renewcommand{\algref}[1]{Alg. \ref{#1}}

\makeatletter
 \def\hlinewd#1{%
      \noalign{\ifnum0=`}\fi\hrule \@height #1 \futurelet
      \reserved@a\@xhline}

\newcommand{\cmark}{\ding{51}}%
\newcommand{\xmark}{\ding{55}}%

\definecolor{red}{rgb}{1,0,0}

\definecolor{blue}{rgb}{0,0,1}

\definecolor{green}{rgb}{0,0.5,0}

\title{Self-balanced Learning for Domain Generalization}
\name{Jin Kim$^\dag$\quad Jiyoung Lee$^\dag$\quad Jungin Park$^\dag$\quad Dongbo Min$^\ddag$ \quad Kwanghoon Sohn$^\dag$\textsuperscript{,*}\thanks{$^{*}$Corresponding author}}
\address{$^\dag$School of Electrical and Electronic Engineering, Yonsei University, Seoul, Korea\\
$^\ddag$Department of Computer Science and Engineering, Ewha Womans University, Seoul, Korea\\
E-mail: khsohn@yonsei.ac.kr}
\begin{document}
%\ninept
%
\maketitle

\begin{abstract}
    Domain generalization aims to learn a prediction model on multi-domain source data such that the model can generalize to a target domain with unknown statistics.
    Most existing approaches have been developed under the assumption that the source data is well-balanced in terms of both domain and class.
    However, real-world training data collected with different composition biases often exhibits severe distribution gaps for domain and class, leading to substantial performance degradation.
    In this paper, we propose a self-balanced domain generalization framework that adaptively learns the weights of losses to alleviate the bias caused by different distributions of the multi-domain source data.
    The self-balanced scheme is based on an auxiliary reweighting network that iteratively updates the weight of loss conditioned on the domain and class information by leveraging balanced meta data.
    Experimental results demonstrate the effectiveness of our method overwhelming state-of-the-art works for domain generalization.
\end{abstract}

\begin{keywords}
Domain generalization, domain imbalance, class imbalance, meta-learning
\end{keywords}

\let\thefootnote\relax\footnote{This work was supported by Institute of Information communications Technology Planning \& Evaluation (IITP) grand funded by the Korea government(MSIT) (No.2020-0-00056, To create AI systems that act appropriately and effectively in novel situations that occur in open worlds.)}
\section{Introduction}
\label{sec:intro}
    Despite the impressive advent of deep learning technologies, the domain shift is still a huge obstacle for deploying them in numerous computer vision tasks.
    As most approaches are trained under the assumption that training and test data are sampled from the same distribution, they often fail to infer accurate predictions for unseen test data sampled from out-of-distribution~\cite{RSC,domainshift,ganin2015unsupervised,ben2007analysis,sun2016return}.

    To overcome this issue, the domain generalization (DG) approaches have attempted to design a generalizable model to minimize the domain gap between train and test domains and perform evenly well under multiple data distributions.
    As part of this effort, the data augmentation methods have been proposed to synthesize the source domains to a wider span of the data space.
    Volpi~\etal~\cite{volpi2018generalizing} adversarially generated fictitious samples by defining target distributions within a certain Wasserstein distance.
    Recently, Yue~\etal~\cite{yue2019domain} employed style transfer~\cite{zhu2017unpaired} to synthesize new training samples, and Carlucci~\etal~\cite{jigsaw} proposed to solve a jigsaw puzzle by shuffling image patches to improve the generalization performance of image classification.
    Although these approaches have demonstrated promising results with data augmentation, it is fairly hard to generate new training samples that can fully cover the real-world distribution, often leading to overfitting issues.
    
    \begin{figure}[t]
    \centering
    	\renewcommand{\thesubfigure}{}%\hfill
    	\subfigure[(a) Target domain: \textbf{C}altech]{\includegraphics[width=1\linewidth]{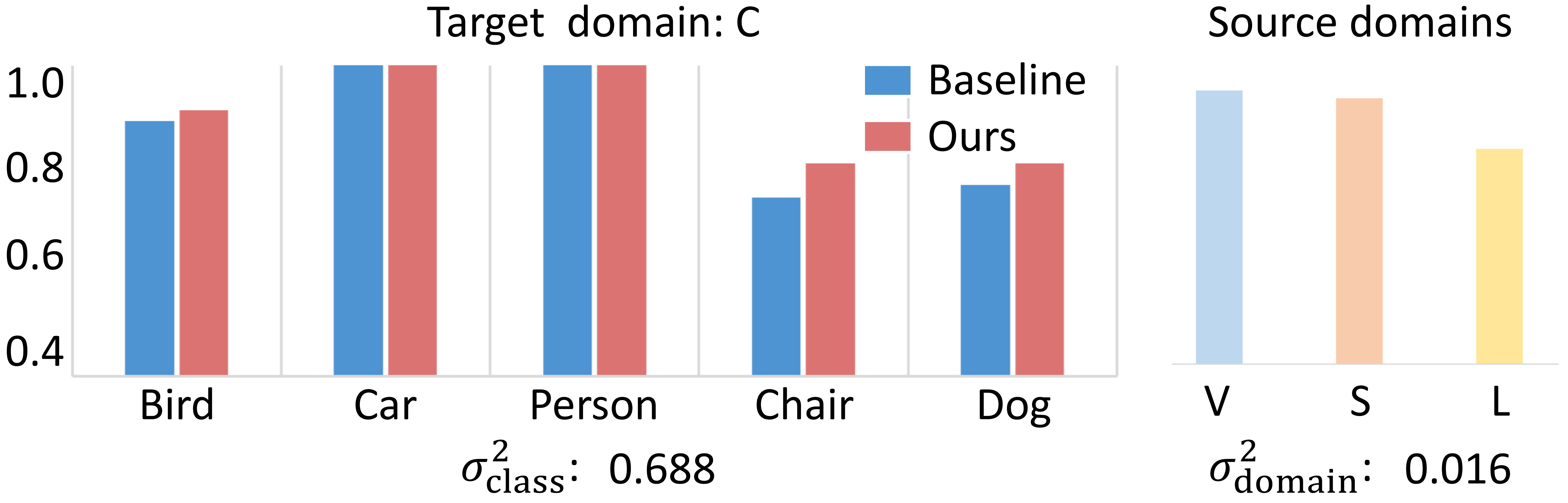}}\hfill \\

    	\subfigure[(b) Target domain: \textbf{L}abelMe]{\includegraphics[width=1\linewidth]{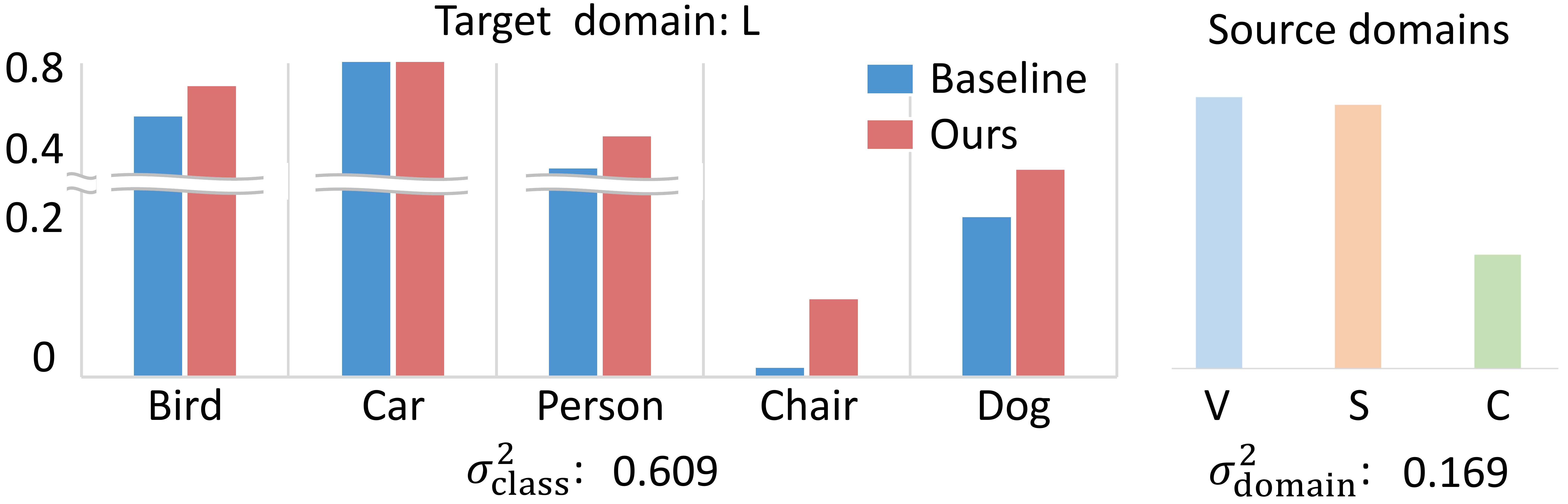}}\hfill

    	\caption{
    	Left: the classification result of each class on the VLCS dataset.
    	Right: the number of samples in source domains.
        The variances of samples with respect to classes and domains, $\sigma^2_\text{class}$ and $\sigma^2_\text{domain}$, are reported, where a large variance value implies severe imbalance of training data.
    }\label{fig:1}
    \end{figure} 
    
    \begin{figure*}[t]
    	\centering
        {\includegraphics[width = 0.95\linewidth]{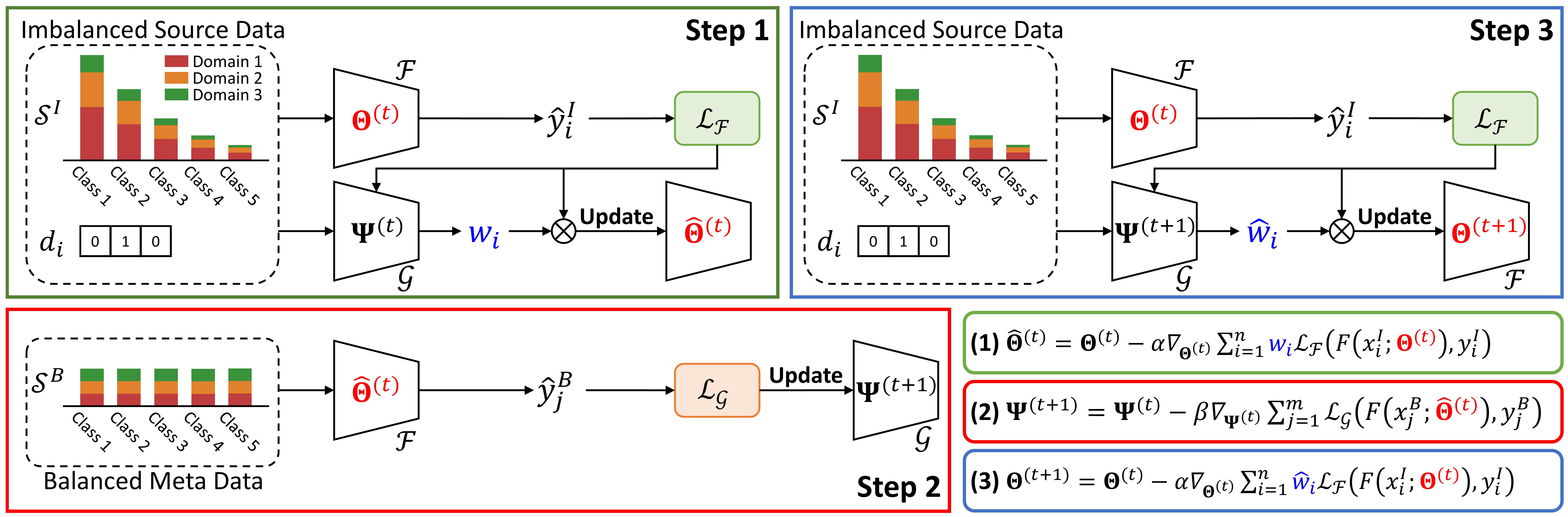}}
        \vspace{-10pt}
     	\caption{
     	An overall methodological flow of SBDG. In step 1, the parameter $\mathbf{\Theta}^{(t)}$ of the task network $\mathcal{F}$ is updated to $\hat{\mathbf{\Theta}}^{(t)}$ with a mini-batch from the imbalanced set.
     	The parameter $\mathbf{\Psi}^{(t)}$ of the reweighting network $\mathcal{G}$ is updated to $\mathbf{\Psi}^{(t+1)}$ with a mini-batch from the balanced set in step 2.
     	As a final step, $\mathbf{\Theta}^{(t)}$ is optimized to $\mathbf{\Theta}^{(t+1)}$ with the imbalanced set and the adaptive weight $\hat{w}_i$ derived from $\mathcal{G}(|{\mathbf{\Psi}^{(t+1)}})$.
     	}\label{fig:2}\vspace{-9pt}
     \end{figure*}
    
    Learning domain invariant feature representations has become an attractive alternative to make the model robust to the unseen target data~\cite{mmd-aae, epifcr, seo2020learning, matsuura2020domain,motiian2017unified}.
    As a pioneering work, maximum mean discrepancy (MMD) constraint~\cite{mmd} was applied to adversarial autoencoders to measure the alignment of the distribution among multiple source domains~\cite{mmd-aae}.
    Some approaches have improved generalization ability through a contrastive loss to embed training samples nearby latent space~\cite{motiian2017unified} or an episodic training procedure to simulate domain shift in training phase~\cite{epifcr}.
    The model-agnostic meta-learning approaches~\cite{finn2017model,metareg,mldg} have been introduced to learn a way that improves the domain generality of a base learner by finding a route to optimization.
    While they have achieved remarkable advances in domain generalization, biased training data distributions in terms of domains and classes behave as the main obstacle impeding the aforementioned approaches from achieving higher accuracy.
    For example, we depict the classification performance on the different target domain using the representative meta-learning based DG method~\cite{mldg} and our proposed method in \figref{fig:1}.
    The variances for the number of data with respect to domains and classes are also reported.
    The accuracy has been degraded significantly from the imbalanced source domains, when evaluating the model on the L dataset in \figref{fig:1}(b).
    \vspace{-2pt}
    
    In this paper, we propose a novel self-balanced domain generalization framework, termed SBDG, that acts as a more effective alternative for the domain/class imbalance problem by explicitly weighting a training loss.
    To this end, we extend the sample reweighting scheme~\cite{metaweight}, which was originally proposed for solving the class imbalance issue, to prioritize the minority domain with relatively higher training losses.
    Specifically, we formulate an auxiliary reweighting network that can be integrated with the domain generalization methods through the adaptive balancing of the training loss to fully leverage the information of a given domain.
    Furthermore, model-agnostic meta-learning~\cite{finn2017model} is employed to train the reweighting network guided by an unbiased meta-dataset that is uniformly distributed in terms of domains and classes as shown in \figref{fig:2}.
    Experimental results show that our framework helps to prevent the performance degeneration by imbalanced training samples and achieves state-of-the-art performance compared to prior works. 
    \vspace{-1pt}
\section{Proposed Method}

\subsection{Problem Statement}
    The objective of domain generalization is to learn a system to perform well in an unseen domain $\mathcal{T}$ using a set of observable source domains $\mathcal{S} = \lbrace D^{k} \rbrace_{k=1}^{K}$, where $K$ is the number of source domains.
    The $k^{th}$ source domain for $c^{th}$ class contains $N^{k}_c$ pairs of input images and class labels $\lbrace x_i^{k}, y_i^{k} \rbrace_{i=1}^{N^{k}_c}$, where $N^{k}_c$ largely varies.
    The drawback of most existing methods is to give equal weight to sample in training without consideration of the imbalance.
    
    Our method improves generalization performance by estimating adaptive learning weights between dominant and minor samples in each training iteration.
    As shown in \figref{fig:2}, the proposed networks are composed of two parts; (1) the task network to perform a target task and (2) the auxiliary reweighting network to balance the task loss of each sample.
    The task network $\mathcal{F}$ predicts the probability of the image class from an input image $x_{i}^{k}$ with a parameter $\mathbf{\Theta}$.
    The auxiliary reweighting network $\mathcal{G}$ takes a loss and the conditional domain vector $d_i$ as inputs and outputs an adaptive weight $w_i$ for $i^{th}$ sample with a network parameter $\mathbf{\Psi}$.
    Although the image classification is employed as the target task in this work, our framework can also be extended in numerous computer vision tasks such as object detection~\cite{gan2016learning} and semantic segmentation~\cite{zhao2019multi}. We remain this as future work.
    
    \subsection{Training}\label{training}
    
    To jointly train the reweighting and task networks, we first divide the source domains into balanced meta dataset $\mathcal{S}^B$ and imbalanced set $\mathcal{S}^I$:
    \begin{equation}
        \mathcal{S} = \mathcal{S}^B \cup \mathcal{S}^I.
    \end{equation}
    The images in the balanced meta dataset $\mathcal{S}^B$ are uniformly distributed in terms of domains and classes.
    The overall framework is illustrated in \figref{fig:2}.
    We take two state-of-the-arts (MLDG~\cite{mldg} or RSC~\cite{RSC}) as the task network $\mathcal{F}$ for the image classification, but any kind of DG models can also be adopted.
    The auxiliary reweighting network $\mathcal{G}$ is composed of 2 fully connected layers with the ReLU activation function, and the sigmoid output layer is adopted to ensure $w_i$ $\in$ $[0,1]$.
    Next, we describe the three training steps in detail.

    \vspace{3pt} \noindent \textbf{Step 1.}
    In the first step, we update the current parameter of the task network with the imbalanced set.
    The image class is predicted by $\mathcal{F}$ with softmax function as follows:
    \begin{equation}\label{eq:softmax}
        \hat{y}^I_i = softmax(\mathcal{F}(x^I_i|\mathbf{\Theta}^{(t)})),
    \end{equation}
    where $\hat{y}^I_i \in \mathbb{R}^{C}$ is the probability of image classes, $x_{i}^{I}$ is a training image sampled from imbalanced set $\mathcal{S}^{I}$, and $C$ is the number of class.
    The task loss $\mathcal{L}_{\mathcal{F}}$ is cross-entropy loss between $\hat{y}^I_i$ and $y^I_i$. While the task loss $\mathcal{L}_{\mathcal{F}}$ can be defined in a different form depending on $\mathcal{F}$, we just use the cross-entropy loss function for simplification.
    Different from~\cite{metaweight} that uses only loss of the task network, the reweighting network $\mathcal{G}$ takes the task loss $\mathcal{L}_{\mathcal{F}}$ and the one-hot domain conditional vector $d_i$ to ensure that $\mathcal{G}$ recognizes the characteristics of the domain.
    Thus, the weight $w_i$ for loss is estimated as follows:
    \begin{equation}\label{eq:initweight}
        w_i = \mathcal{G}(\Pi(d_i,\mathcal{L}_{\mathcal{F}})|\mathbf{\Psi}^{(t)}),
    \end{equation}
    where $\Pi(\cdot)$ represents a concatenation operation.
    The current parameter $\mathbf{\Theta^{(t)}}$ of the task network is optimized by minimizing the following weighted loss:
    \begin{equation}\label{eq:thetainner}
        \hat{\mathbf{\Theta}}^{(t)} = \underset{\mathbf{\Theta}^{(t)}}{\operatorname{argmin}}\sum_{i=1}^{n}{w_i \mathcal{L}_{\mathcal{F}}(\hat{y}_{i}^{I}, y_{i}^{I})},
    \end{equation}
    where $n$ is the number of samples in a mini-batch.

    \vspace{3pt} \noindent \textbf{Step 2.}
    The parameter $\mathbf{\Psi}^{(t)}$ of the reweighting network is optimized to be guided by the parameter $\hat{\mathbf{\Theta}}^{(t)}$ on the balanced meta dataset $\mathcal{S}^{B}$ by optimizing the following equation:
    \begin{equation}\label{eq:psiupdate}
        \mathbf{\Psi}^{(t+1)} = \underset{\mathbf{\Psi}^{(t)}}{\operatorname{argmin}}\sum_{j=1}^{m}{\mathcal{L}_{\mathcal{G}}(\hat{y}_{j}^{B}, y_{j}^{B})},
    \end{equation}
    where $\mathcal{L}_{\mathcal{G}}$ is a cross entropy loss, and $x_{j}^{B}$ is a training image sampled from the balanced meta dataset $\mathcal{S}^{B}$, $\hat{y}_{j}^{B}$ is the estimated class vector with the parameter $\hat{\mathbf{\Theta}}^{(t)}$, and $m$ is the size of mini-batch.

    \begin{algorithm}[t]
     \caption{Self-Balanced Domain Generalization}
     \textbf{Hyperparameters}: Mini-batch size $n$,$m$; max iteration $T$; step size $\alpha$, $\beta$ \\
     \textbf{Input}: Source training domains $\mathcal{S}$; imbalanced training set $\mathcal{S}^{I}$; balanced training set $\mathcal{S}^{B}$ \\
     \textbf{Output}: Task parameter $\mathbf{\Theta}^{(T)}$
     \label{alg:1}
     \begin{algorithmic}[1]
     \Procedure{Training}{$\mathcal{S}^{I}$,$\mathcal{S}^{B}$}
      \State Initialize parameters $\mathbf{\Theta}$ and $\mathbf{\Psi}$
      \For { $t = 0$ \textbf{to} $T-1$}
        \State {$\{x^I_i, y^I_i,d_i\} \leftarrow SampleMiniBatch(\mathcal{S}^{I},n).$}
        \State {$\{x^B_j, y^B_j\} \leftarrow SampleMiniBatch(\mathcal{S}^{B}.m).$}
        \State {$\hat{y}^I_i = softmax(\mathcal{F}(x^I_i|\mathbf{\Theta}^{(t)})).$}
        \State {$w_i = \mathcal{G}(\Pi(d_i,\mathcal{L}_{\mathcal{F}})|\mathbf{\Psi}^{(t)}).$}
        \State{$\hat{\mathbf{\Theta}}^{(t)} \leftarrow \mathbf{\Theta}^{(t)} - \alpha \frac{1}{n} \sum_{i=1}^{n}{w_i \bigtriangledown_{\mathbf{\Theta}} \mathcal{L}_{\mathcal{F}}(\hat{y}^I_i, y^I_i)}.$}
        \State {$\hat{y}^B_j = softmax(\mathcal{F}(x^B_j|\hat{\mathbf{\Theta}}^{(t)})).$}
        \State {$\mathbf{\Psi}^{(t+1)} \leftarrow \mathbf{\Psi}^{(t)} - \beta \frac{1}{m} \sum_{j=1}^{m}{\bigtriangledown_{\mathbf{\Psi}}\mathcal{L}_{\mathcal{G}}(\hat{y}_{j}^{B}, y_{j}^{B})}.$}
        \State {$\hat{w}_i = \mathcal{G}(\Pi(d_i,\mathcal{L}_{\mathcal{F}})|\mathbf{\Psi}^{(t+1)}).$}
        \State {$\mathbf{\Theta}^{(t+1)} \leftarrow \mathbf{\Theta}^{(t)} - \alpha \frac{1}{n} \sum_{i=1}^{n}{\hat{w}_i \bigtriangledown_{\mathbf{\Theta}} \mathcal{L}_{\mathcal{F}}(\hat{y}^I_i, y^I_i)}.$}
     \EndFor
     \EndProcedure
     \end{algorithmic}  
    \end{algorithm}
    
    \vspace{3pt} \noindent \textbf{Step 3.}
    At the last step, the updated parameter $\mathbf{\Psi}^{(t+1)}$ is used to produce the adaptive weight $\hat{w}_i$ and the parameter $\mathbf{\Theta}^{(t)}$ is optimized as follows:
    \begin{equation}\label{eq:thetaupdate}
        \mathbf{\Theta}^{(t+1)} = \underset{\mathbf{\Theta}^{(t)}}{\operatorname{argmin}}\sum_{i=1}^{n}{\hat{w}_i \mathcal{L}_{\mathcal{F}}(\hat{y}_{i}^{I}, y_{i}^{I})},
    \end{equation}
    where $\hat{w}_i = \mathcal{G}(\Pi(d_i,\mathcal{L}_{\mathcal{F}})|\mathbf{\Psi}^{(t+1)})$ and $\hat{y}_{i}^{I}$ is derived from the task network with the parameter $\mathbf{\Theta}^{(t)}$.
    The flow of SBDG algorithm is summarized in \algref{alg:1} and the convergence of this second gradient update procedure can be mathematically proven in a similar way to \cite{metaweight}.

\label{sec:method}
\section{Experiments}
\label{sec:experiments}

    \begin{figure*}[!t]
        \centering
        	\renewcommand{\thesubfigure}{}%\hfill
        	\subfigure[(a)]{\includegraphics[width=0.247\linewidth]{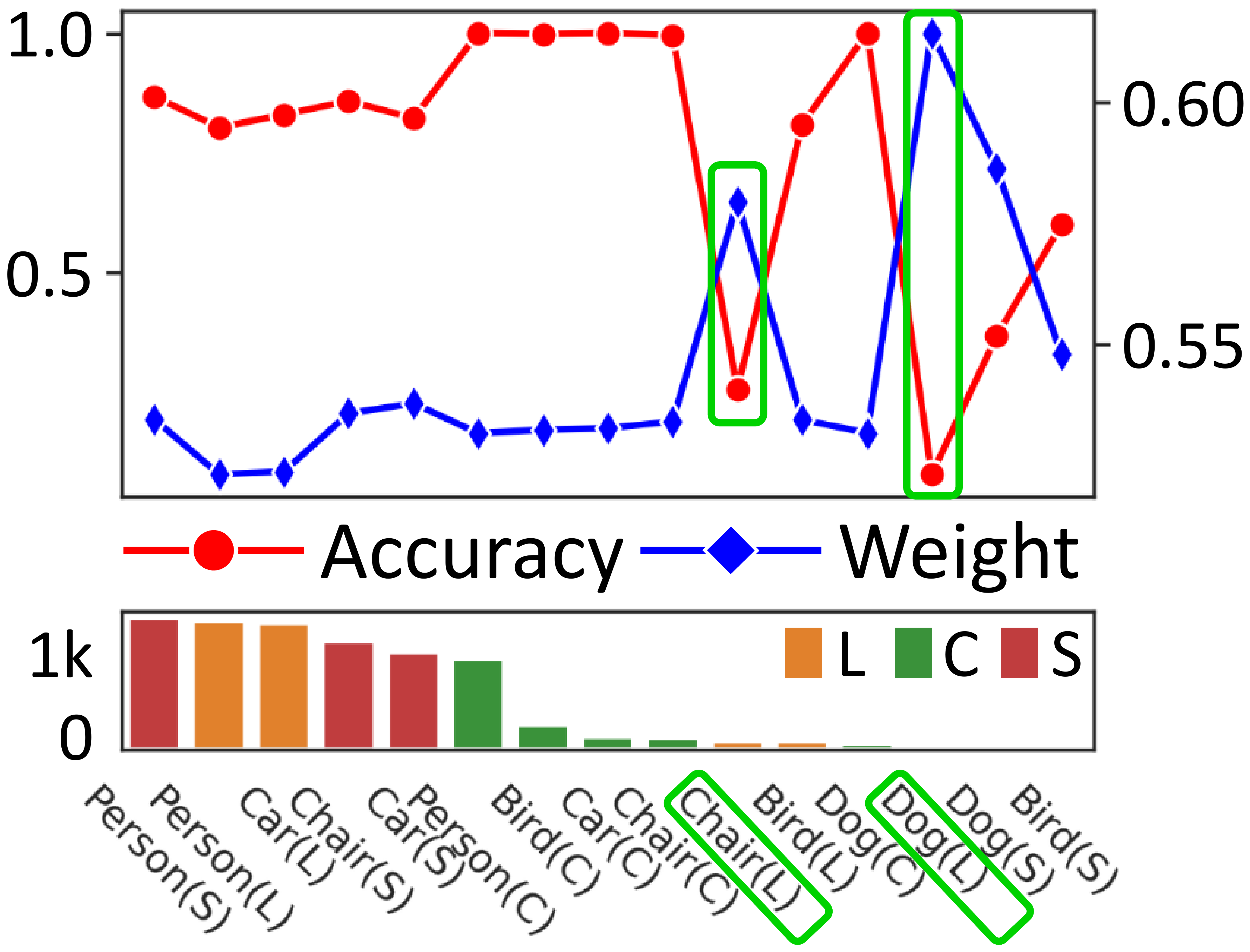}}\hfill
        	\subfigure[(b)]{\includegraphics[width=0.248\linewidth]{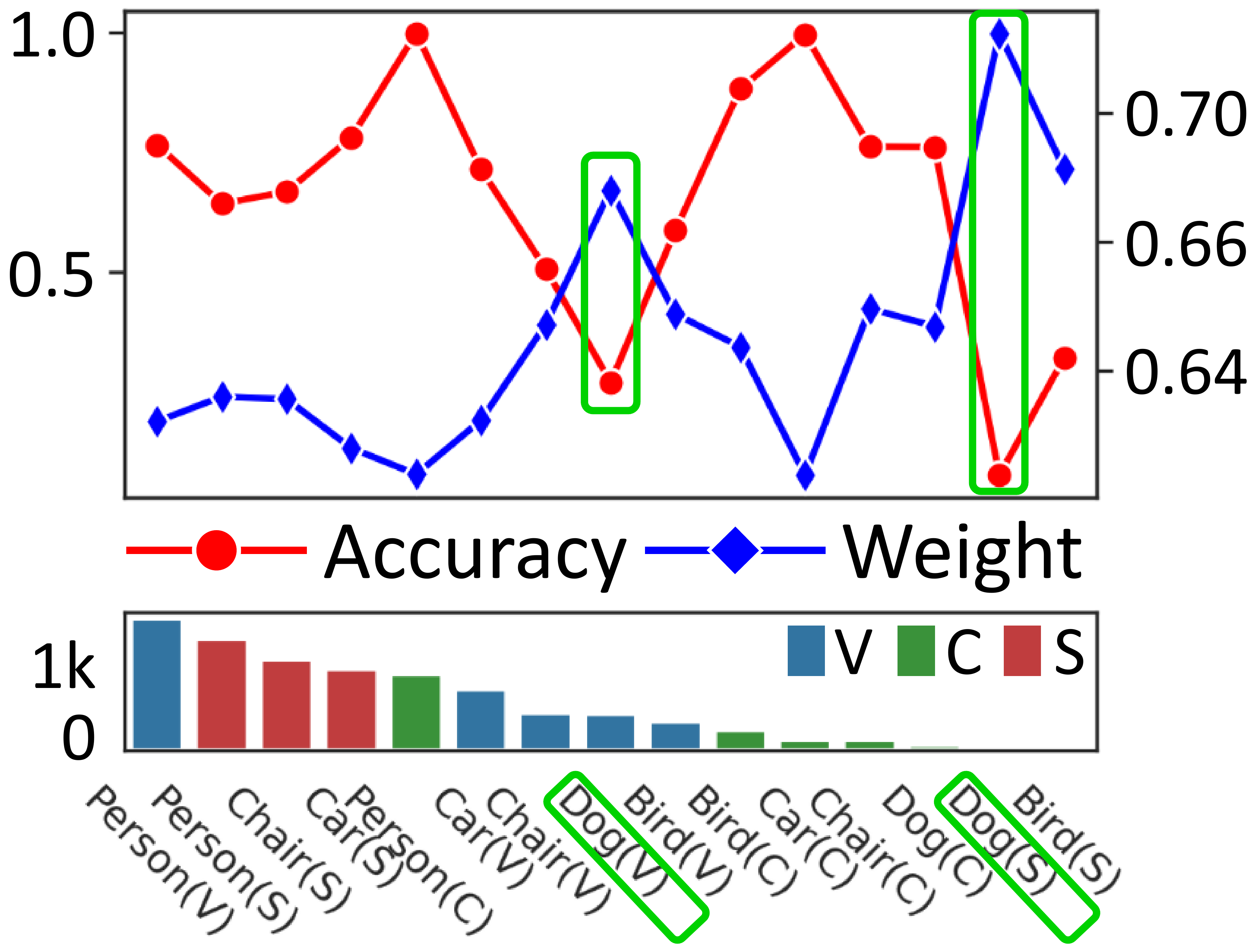}}\hfill
        	\subfigure[(c)]{\includegraphics[width=0.236\linewidth]{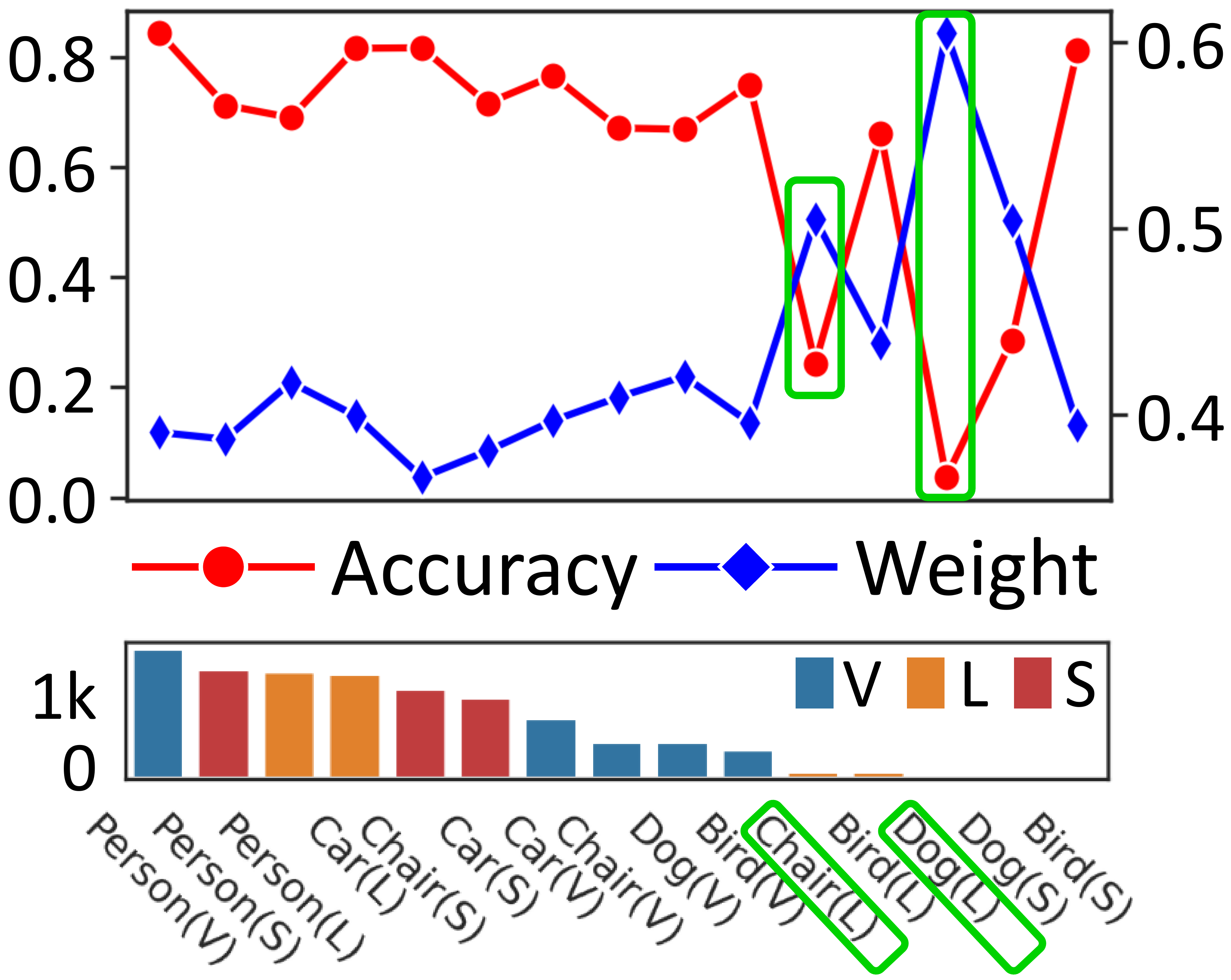}}\hfill
        	\subfigure[(d)]{\includegraphics[width=0.25\linewidth]{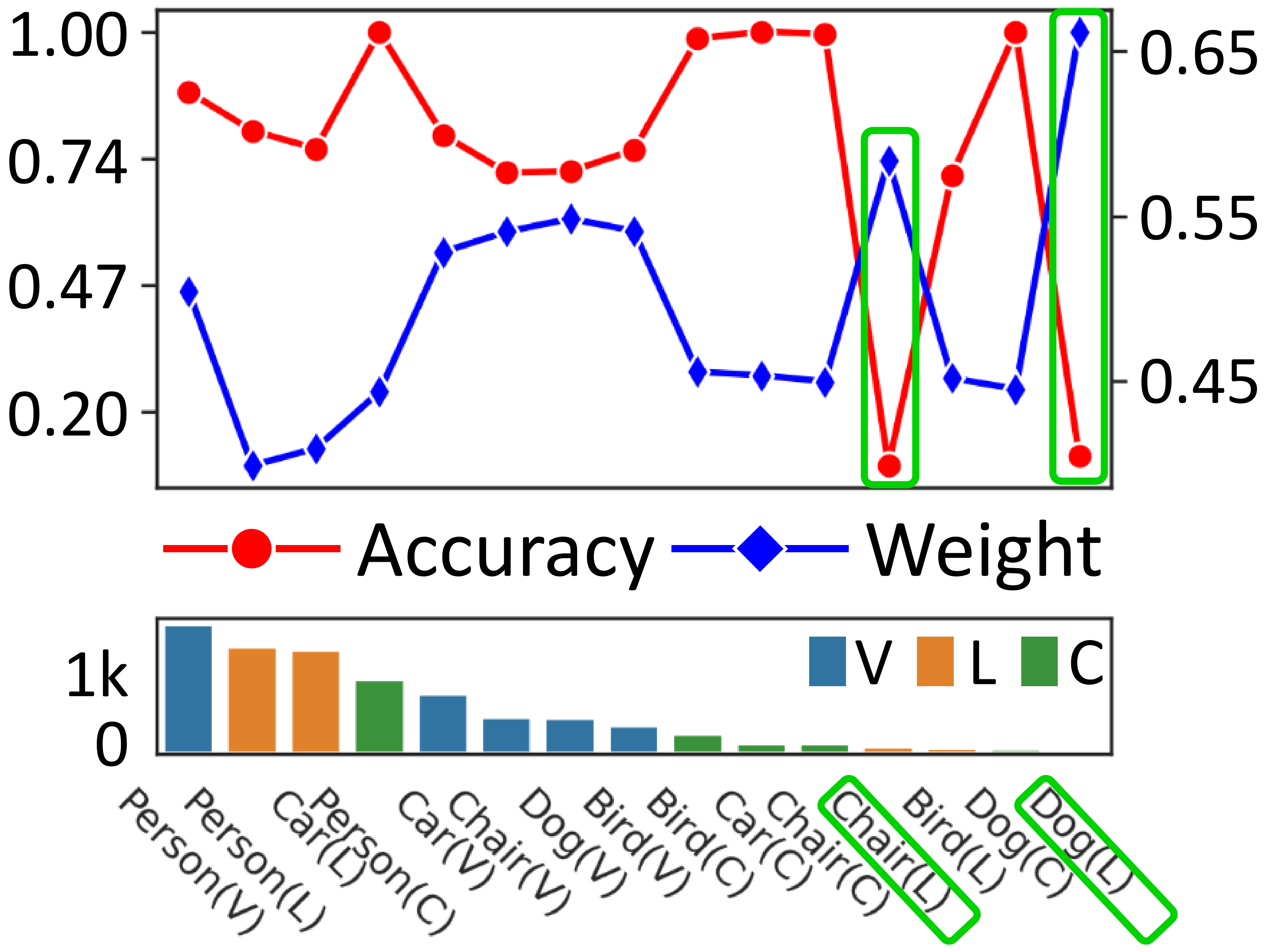}}\hfill
        	\vspace{-14pt}
        	\caption{
        	The accuracy (red line) and the adaptive weight (blue line) are depicted in the top row, and the number of images is shown in the bottom row for each class in three source domains on VLCS~\cite{vlcs} dataset.
            The target domain is set to (a) V, (b) L, (c) C, and (d) S, respectively.
            We measure the accuracy before training step 1 and the weight at training step 3 (see \secref{training}). 
        }\label{fig:3}
        \vspace{-20pt}
    \end{figure*}

\vspace{-3pt}\subsection{Experimental Settings}

    \vspace{-3pt}\noindent\textbf{Dataset.}
    We evaluate our method on VLCS dataset~\cite{vlcs}, which is the commonly used domain generalization benchmark for image classification. It contains images from four different datasets (domains): VOC2007 (V), LabelMe (L), Caltech (C), and SUN09 (S). Each domain includes five classes: bird, car, chair, dog, and person.
    Following the standard protocol of \cite{vlcsprotocol}, the VLCS dataset is randomly divided for each domain into 70\% training and 30\% test set.

    \vspace{3pt}\noindent\textbf{Implementation details.}
    We picked 12 samples from each domain-class pair set to construct the balanced meta dataset by leveraging random-over-sampling from 30\% of the training set. The mini-batch size of the reweighting network was 9 for each domain, totaling 27 for three source domains. The learning rate of reweighting network $\beta$ is set as $5\times10^{-5}$.
    Following the previous works~\cite{jigsaw,RSC}, we perform random cropping, horizontal flipping, and RGB to greyscale converting.
    
    To evaluate the influence of our framework, two baseline methods were adopted as the task network $\mathcal{F}$; (1) MLDG~\cite{mldg} and (2) RSC~\cite{RSC}. MLDG~\cite{mldg} is the well-known meta-learning based DG algorithm that meta-learns how to generalize across domains.
    RSC~\cite{RSC} is the current state-of-the-art DG algorithm that discards a few percent of high gradient features at each epoch and trains the model with the remaining information.
    To train with MLDG~\cite{mldg}, outer and inner update's gradients are added equally. The reweighted loss was used both in the inner and outer update of MLDG~\cite{mldg}. The batch size is 128 for each source domain and the learning rate $\alpha$ is set as $5\times10^{-4}$.
    The remaining hyper-parameters for RSC were used the same as the original method~\cite{RSC}.
    We used AlexNet pre-trained on ImageNet as a backbone network for each baseline method.
    We selected the best model via leave-one-domain-out validation model~\cite{masf}.

    \begin{table}[t]
    \caption{Performance comparison on VLCS dataset with state-of-the-art methods. $^\dagger$ denotes the combination of SBDG with MLDG~\cite{mldg} as the task network, and $^\ddagger$ denotes that the task network is RSC~\cite{RSC}. $(\cdot)$ denotes the normalized variance of source domains corresponding to the target domain.}
    \vspace{5pt}
        \centering
        \resizebox{1\linewidth}{!}{
            \begin{tabular}{lccccc}
            \hlinewd{0.8pt}
            \multirow{2}{*}{Method}   & \textbf{C}altech101           & \textbf{L}abelMe           & \textbf{S}UN09           & \textbf{V}OC2007           & \multirow{2}{*}{Avg.}         \\
            & (0.016)       & (0.169)       & (0.160)       & (0.150)  \\
            \hline \hline
            Deep-All~\cite{jigsaw}                 & 96.25       & 59.72       & 64.51       & 70.58       & 72.76                 \\
            CIDDG~\cite{ciddg}                 & 88.83       & 63.06       & 62.10       & 64.38       & 69.59                 \\
            Undo-Bias~\cite{undobias}                 & 93.63       & 63.49       & 61.32       & 69.99       & 72.11                 \\
            MMD-AAE~\cite{mmd-aae}                 & 94.40       & 62.60       & 64.40       & 67.70       & 72.28                 \\
            Epi-FCR~\cite{epifcr}                 & 94.10       & 64.30       & 65.90       & 67.10       & 72.90                 \\
            JiGen~\cite{jigsaw}                 & 96.93       & 60.90       & 64.30       & 70.62       & 73.19                 \\
            MASF~\cite{masf}                 & 94.78       & 64.90       & 67.64       & 69.14       & 74.11                 \\
            MLDG~\cite{mldg}                 & 94.40       & 61.30       & 65.90       & 67.70       & 72.30                 \\
            RSC~\cite{RSC}            & \textbf{97.61}       & 61.86       & 68.32       & \textbf{73.93}       & 75.43                 \\ \hline
            Ours$^\dagger$ with MLDG                 & 95.05       & 63.97       & 66.08      & 67.45        & 73.14                 \\
            Ours$^\ddagger$ with RSC            & 96.47       & \textbf{65.85}       & \textbf{68.68}       & 72.59       & \textbf{75.90}                 \\
            \hlinewd{0.8pt}
            \end{tabular}
        }
        \label{tab1}
        \vspace{-14pt}
    \end{table}

\vspace{-9pt}
\subsection{Results}
    \vspace{-3pt}In~\tabref{tab1}, we compared our model with several state-of-the-art models on VLCS dataset~\cite{vlcs} in terms of classification accuracy on different target domains.
    The variation of source domains ($\sigma^2_\text{domain}$) is computed for each target domain, e.g. using `L', `S, and `V' source domains for `C' target domain.
    While most algorithms work reasonably well for `C' domain, they perform poorly when `L', `S', and `V' are target domains.
    The reason is that when the target domain is the `C', $\sigma^2_\text{domain}$ is lower than in other cases as shown in \tabref{tab1}. 
    Our method shows the significant improvement of the accuracy when $\sigma^2_\text{domain}$ is high (\ie~target domain is `L' and `S').
    We achieve 75.90\% accuracy on average when the task network is RSC~\cite{RSC}.
    Especially, the performance gain is 3.99\% in the `L' domain and 0.36\% in the `S' domain over RSC~\cite{RSC}.
    A similar performance gain is also observed when MLDG~\cite{mldg} is employed as the task network.

\vspace{-9pt}
\subsection{Analysis} 
    \vspace{-3pt}\tabref{tab2} shows the effectiveness of the domain condition vector used in the auxiliary reweighting network.
    The domain information assigned to the reweighting network creates more appropriate weights.
    The conditional domain vector makes the reweighting network to be aware that there are other types of imbalances depending on the source domains.
    In \figref{fig:3}, we investigated the adaptive weight (blue line) and the accuracy (red line) for each class corresponding to the source domains at the same iteration (top row) and the number of training data for each class and domain (bottom row).
    Note that, the accuracy is measured before training step 1 and the adaptive weight is measured at training step 3.
    The results show that the domain-class imbalances degrade the performance in some classes and domains (green box in \figref{fig:3}). The weight learned from the proposed reweighting network is inversely proportional to the accuracy, indicating that it emphasizes the loss to learn in a balanced manner. This demonstrates that the proposed method copes well with the domain-class imbalance problem.

    \begin{table}[!t]
        \caption{Ablation study for the effect of the conditional domain vector on VLCS dataset. We use RSC~\cite{RSC} as the task network.}
        \label{tab2}\vspace{5pt}
        \centering
        \resizebox{1\linewidth}{!}{
            \begin{tabular}{ccccccc}
            \hlinewd{0.8pt}
            {Domain vector}   & \textbf{C}altech101           & \textbf{L}abelMe           & \textbf{S}UN09           & \textbf{V}OC2007           & {Avg.}         \\
            \hline
            \xmark            & 95.97       & 62.31       & 67.58       & 71.34       & 74.30                 \\
            \cmark            & 96.47       & 65.85       & 68.68       & 72.59       & 75.90                 \\
            \hlinewd{0.8pt}
            \end{tabular}
        }
        \vspace{-14pt}
    \end{table}
\section{Conclusion}
\vspace{-7pt}In this work, we introduce a novel self-balanced domain generalization method to deal with the imbalanced distribution in terms of domains and classes.
we predict adaptive weights of losses through the auxiliary reweighting network in the training phase to effectively balance the impact of training samples.
Our method outperforms prior approaches especially when the source domains are largely imbalanced.
Furthermore, we hope that this method can offer a new research direction by addressing the bias problem in the domain adaptation.
\label{sec:conclusion}
\section*{Acknowledgement}
\label{sec:ack}
    \vspace{-3pt}This research was supported by the Yonsei University Research Fund of 2021 (2021-22-0001).

% \bibliographystyle{IEEEbib}
% \bibliography{strings,refs}

\end{document}